\documentclass{article}
\usepackage{ijcai18}

\usepackage{times}
\usepackage{xcolor}
\usepackage{soul}
\usepackage[utf8]{inputenc}
\usepackage[small]{caption}

\usepackage{multirow}
\usepackage{verbatim}
\usepackage{graphicx}
\usepackage{subfigure}
\usepackage[switch]{lineno}

\title{Crowd Counting using Deep Recurrent Spatial-Aware Network}

\author{
Lingbo Liu$^1$,
Hongjun Wang$^1$,
Guanbin Li$^1$\thanks{Corresponding author is Guanbin Li.},
Wanli Ouyang$^2$,
Liang Lin$^1$
\\
$^1$ School of Data and Computer Science, Sun Yat-sen University, Guangzhou, China \\
$^2$ School of Electrical and Information Engineering, The University of Sydney, Sydney, Australia\\
\{liulingb,wanghq8\}@mail2.sysu.edu.cn,
liguanbin@mail.sysu.edu.cn, \\
wanli.ouyang@sydney.edu.au, linliang@ieee.org
}

\begin{document}
\maketitle

\begin{abstract}
 Crowd counting from unconstrained scene images is a crucial task in many real-world applications like urban surveillance and management, but it is greatly challenged by the camera’s perspective that causes huge appearance variations in people’s scales and rotations. Conventional methods address such challenges by resorting to fixed multi-scale architectures that are often unable to cover the largely varied scales while ignoring the rotation variations. In this paper, we propose a unified neural network framework, named Deep Recurrent Spatial-Aware Network, which adaptively addresses the two issues in a learnable spatial transform module with a region-wise refinement process. Specifically, our framework incorporates a Recurrent Spatial-Aware Refinement (RSAR) module iteratively conducting two components: i) a Spatial Transformer Network that dynamically locates an attentional region from the crowd density map and transforms it to the suitable scale and rotation for optimal crowd estimation; ii) a Local Refinement Network that refines the density map of the attended region with residual learning.  Extensive experiments on four challenging benchmarks show the effectiveness of our approach. Specifically, comparing with the existing best-performing methods, we achieve an improvement of 12\% on the largest dataset WorldExpo’10 and 22.8\% on the most challenging dataset UCF\_CC\_50.
\end{abstract}

\section{Introduction}

\begin{figure}[t]
\centering
   \includegraphics[width=0.925\columnwidth]{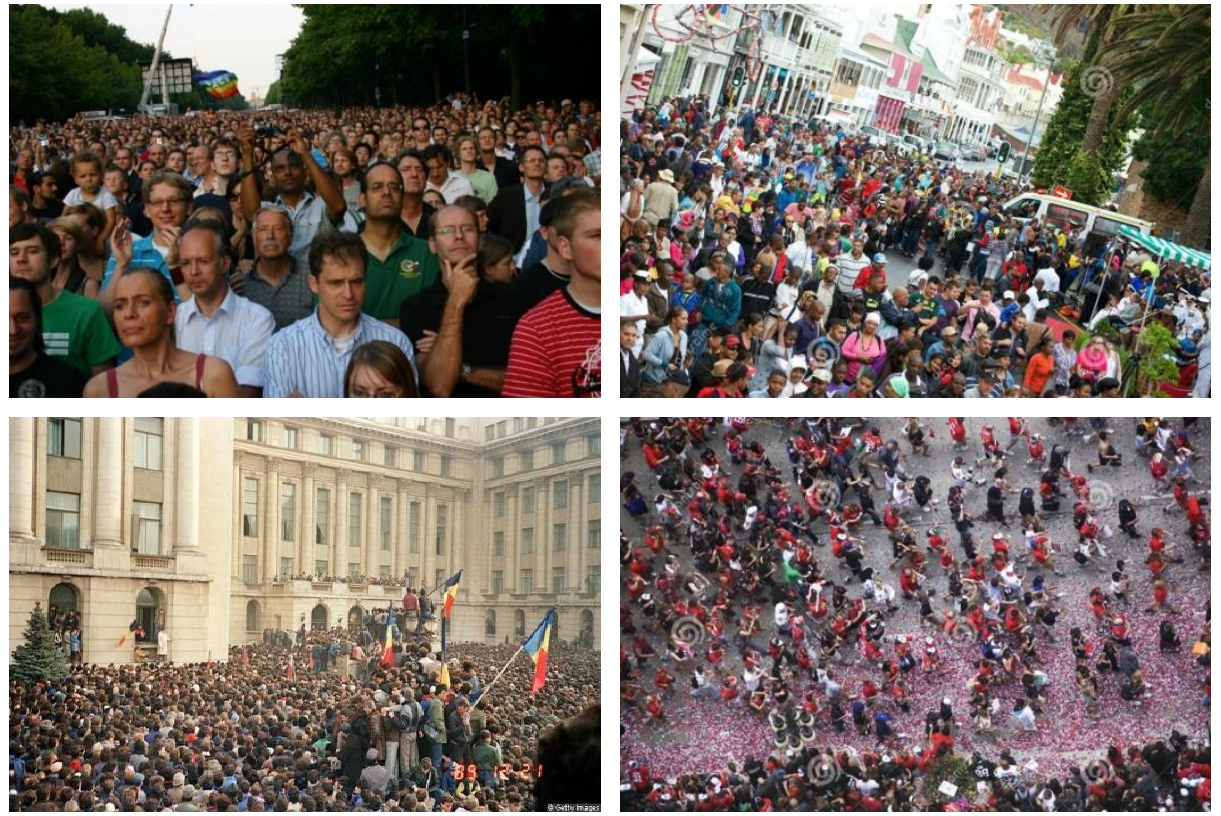}
\vspace{-2mm}
   \caption{Examples of unconstrained crowd scenes from the ShanghaiTech dataset. There are many challenges limiting the performance of crowd counting, i.e, diversity of camera setting, the large variation of scales and rotations of people.
   }
\vspace{-4mm}
\label{fig:challenge}
\end{figure}

Crowd counting, which aims at estimating the total number of people in unconstrained crowded scenes, has received increasing research interests in recent years due to its potential application in many real-world scenarios, such as video surveillance~\cite{xiong2017spatiotemporal} and traffic monitoring~\cite{zhang2017fcn}. 
Despite recent progress, crowd counting remains a challenging issue, especially in the face of extreme occlusions, changes in lighting and camera perspective.

In recent two years, deep convolutional neural networks have been widely used in crowd counting and have made substantial
progress~\cite{sindagi2017cnn,onoro2016towards,sam2017switching,xiong2017spatiotemporal,zhang2015cross}. The success of most existing deep models stems from the effective modeling of the scale variation of people/heads and the varying degrees of regional crowd density. However, all of these methods, without exception, obtain the density estimation of the whole image by merging the prediction results of a number of pre-designed fixed sub-network structures (working on the whole image or image patches). Specifically, they are either proposed to fuses the features from multiple convolutional neural networks with different receptive fields to handle the scale variation of people groups~\cite{zhang2016single}, or directly divide crowd scene into multiple non-overlapping patches and provide a pool of regression networks for each patch selection~\cite{sam2017switching}. Although these strategies can, to a certain extent, improve the adaptability of the prediction to images of diverse density regions or people of various scales, the limited enumeration of a fixed number of tailor-designed networks or several sizes of receptive fields can not well cope with all the scenarios after all. Most importantly, none of these algorithms take into account the impact of different pose and photographic angles on crowd density estimation while crafting their network structures. As shown in Figure~\ref{fig:challenge}, the camera viewpoints in various scenes create different perspective effects and may result in large variation of scales, in-plane and out-plane rotation of people.

To address the aforementioned concerns, we propose a Deep Recurrent Spatial-Aware Network for crowd counting. The core of our network is a Recurrent Spatial-Aware Refinement (RSAR) module, which recurrently conducting refinement on an initial crowd density map through adaptive region selection and residual learning. Specifically, the RSAR module consists of two alternately performed components: i) a Spatial Transformer Network is incorporated in each LSTM step for simultaneously region cropping and warping, which allows the network to adaptively cope with the various degrees of congestion, people scale and rotation variation in the same scene; ii) a Local Refinement Network refines the
density map of the selected region.
In general, the main contributions of this work are three-fold.
\begin{itemize}
\item We provide an adaptive mode to simultaneously handle the effect of both scale and rotation variation by introducing a spatial transform module for crowd counting. To the best of our knowledge, we are the first to address the issue of the rotation variation on this task.
\item We propose a novel deep recurrent spatial-aware network framework to recurrently select a region~(with learnable scale and rotation parameters) from an initial density map for refinement, dependent on feature warping and residual learning.
 \item Extensive experiments and evaluations on several public benchmarks show that our proposed method achieves superior performance in comparison to other state-of-the-art methods. Specifically, compared with the existing best-performing methods, we achieve an improvement of 12\% on the WorldExpo'10 dataset and 22.8\% on the most challenging UCF\_CC\_50 dataset.
\end{itemize}

\section{Related Work}
\textbf{Deep learning Methods for Crowd Counting:}
Inspired by the significant progress of deep learning on various computer vision tasks\cite{zhu2017learning,chen2017recurrent,li2017face,chen2016disc,li2017instance}), many researchers also have attempted to adapt the deep neural network to the task of crowd counting and achieved great success.
Most of the existing methods addressed the scale variation of people with multi-scale architectures.
Boominathan et al.~\cite{boominathan2016crowdnet} proposed to tackle the issue of scale variation using a combination of shallow and deep networks along with an extensive data augmentation by sampling patches from multi-scale image representations.
HydraCNN~\cite{onoro2016towards} proposed to learn a non-linear regression model which used a pyramid of image patches extracted at multiple scales to perform the final density prediction.
A pioneering work was proposed by Zhang et al.~\cite{zhang2016single}, in which they utilized multi-column convolutional neural networks with different receptive fields to learn scale-robust features.
Sam et al.~\cite{sam2017switching} proposed a Switching-CNN to map a patch from the crowd scene to one of the three regression networks, each of which can handle the particular range of scale variation.
CP-CNN~\cite{sindagi2017generating} proposed a Contextual Pyramid CNN to generate high-quality crowd density estimation by incorporating global and local contextual information into the multi-column networks.
However, the above-mentioned methods have two significant limitations.
First, as a neural network with the fixed static receptive field can only handle a limited enumeration of scale variation, these methods do not scale well to large scale change and cannot well cope with all level of scale variation in diverse scenarios.
Second, they did not take the rotation variation of people into consideration, which limits the models' robustness towards camera perspective variations.
To the best of our knowledge, we are the first work to simultaneously address the issue of large scale and rotation variations in an adaptively learnable mode on this task.

\textbf{Deep learning Methods for Crowd Counting:}
Spatial transformer Network (STN) ~\cite{jaderberg2015spatial} is a sub-differentiable sampling-based module, which is designed to spatially transform its input map to an output map that corresponds to a subregion of the input map and can be hence regarded as an effective region selection mechanism.
It is convenient to incorporate a spatial transformer layer to the convolutional neural network and train it with the standard back-propagation algorithm.
A parameter matrix is used to determine the location of the subregion, as well as its resize scale and the rotation angle.
Recently, the spatial transformer has been applied to various computer vision tasks, e.g., multi-label image recognition~\cite{wang2017multi} and saliency detection~\cite{kuen2016recurrent}.
To the best of our knowledge, our work is the first in successfully using multiple iterations of STN within a LSTM framework for crowd counting.

\begin{figure*}[t]
\begin{center}
 \includegraphics[width=1.8\columnwidth]{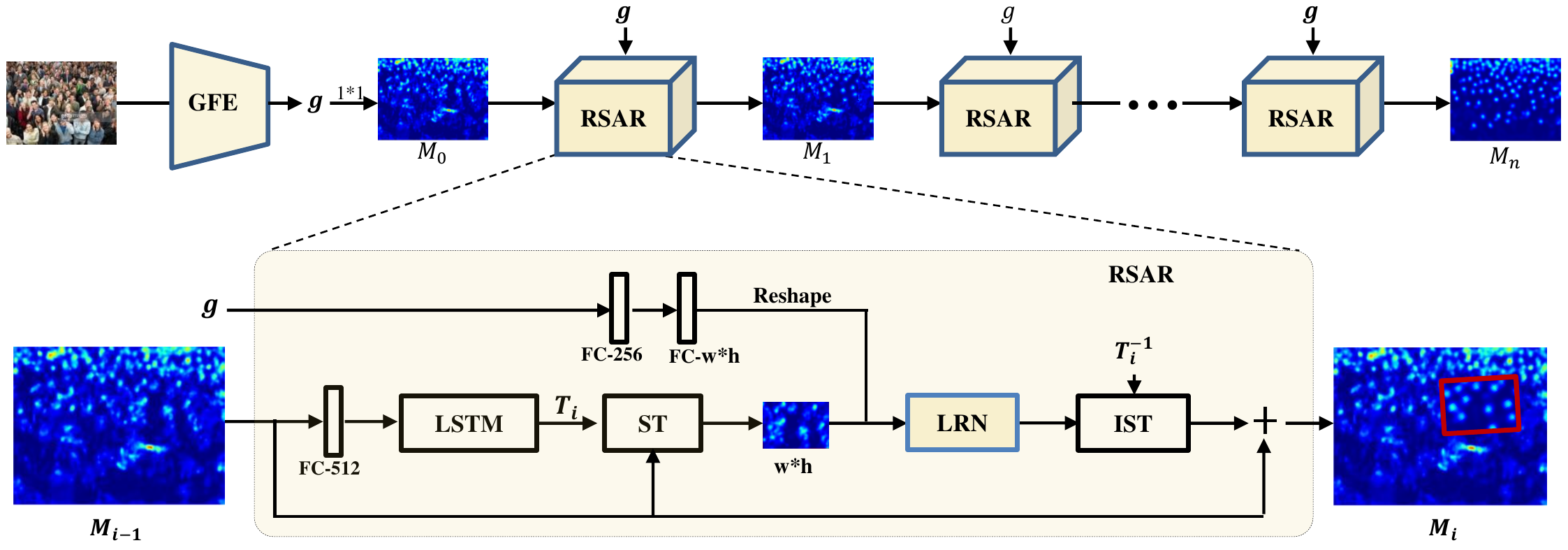}
 \vspace{-5mm}
\end{center}
   \caption{The architecture of the proposed Deep Recurrent Spatial-Aware Network for crowd counting. The Global Feature Embedding (GFE) module takes the whole image to extract its global feature ${g}$, which is used to estimate the initial crowd density map ${M_0}$ using a convolutional layer with a kernel size of ${1\times1}$. The Recurrent Spatial-Aware Refinement (RSAR) module iteratively selects an image region using spatial transformer network and refines the density map of the selected region incorporating global context with the Local Refinement Network (LRN).
   ST denotes the spatial transformer network and IST refers to the inverse spatial transformer network. FC-$N$ stands for the fully-connected layer with $N$ output neurons. }
\vspace{-4mm}
\label{fig:network-structure}
\end{figure*}

\section{Proposed Method}
As illustrated in Figure~\ref{fig:network-structure}, we propose a novel Deep Recurrent Spatial-Aware Network for crowd counting, which is composed of two modules, including a Global Feature Embedding (GFE) module and a Recurrent Spatial-Aware Refinement (RSAR) module.
Specifically, the GFE module takes the whole image as input for global feature extraction, which is further used to estimate an initial crowd density map. And then the RSAR module is applied to iteratively locate image regions with a spatial transformer-based attention mechanism and refine the attended density map region with residual learning. For convenience in the following, we denote the crowd density map in the $i$-th iteration as ${M_i}$. Noted that ${M_0}$ is the initial crowd density map.

\subsection{Global Feature Embedding}~\label{sec:coarse_map_generation}
The goal of the Global Feature Embedding module is to transform the input image into high-dimensional feature maps, which is further used to generate an initial crowd density map of the image.
Inspired by previous works~\cite{zhang2016single,sindagi2017generating}, we develop our GFM module in a multi-scale mode.
As shown in Figure~\ref{fig:mcnn}(a), the GFE module is composed of three columns of CNNs, each of which has seven convolutional layers with different kernel sizes and channel numbers as well as three max-pooling layers.
Given an image ${I}$, we extract its global feature $g$ by feeding it into GFM and concatenating the outputs of all the columns.
After obtaining the global feature $g$, we generate the initial crowd density map ${M_0}$ of image ${I}$ using a convolutional layer with a kernel size of ${1\times1}$, which can be expressed as:
\begin{equation}
g=GFM(I),
M_0 = Conv(g),
\end{equation}
where ${Conv}$ denotes the convolution operation. As shown in Figure~\ref{fig:result_visual}, the initial crowd density is too rough to estimate the number of people in the image. To further improve the quality of crowd density map, we propose a Recurrent Attentive Refinement (RSAR) module to iteratively refine the density map, which will be described in the next subsection.

\subsection{Recurrent Spatial-Aware Refinement}
In this section, we propose a novel Recurrent Attentive Refinement (RSAR) module to iteratively refine the crowd density map.
Our proposed RSAR consists of two alternately performed components: i) a Spatial Transformer Network dynamically locates an attentional region from the crowd density map; ii) a Local Refinement Network refines the density map of the selected region with residual learning.
A high-quality crowd density map with accurately estimated crowd number would be acquired after a refinement of ${n}$ iterations.

\subsubsection{Attentional Region Localization}
In the $i$-th iteration, we first determine which image region to be refined with a spatial transformer network. We encode ${M_{i-1}}$ into a 512-dimensions feature with a fully-connected layer and put it into a Long Short-Term Memory (LSTM)~\cite{hochreiter1997long} layer, which can be formulated as:
\begin{equation}
c_i, h_i = \textbf{LSTM}(c_{i-1}, h_{i-1}, FC(M_{i-1})),
\end{equation}
where ${c_i, h_i}$ are the memory cell and hidden state of current iteration and FC is the fully-connected layer.
The LSTM is utilized for capturing the past information of the density maps and transformation actions.
We define the transformation matrix ${T_i}$ of the spatial transformer as:
\begin{equation}
T_i={
\left[ \begin{array}{ccc}
\theta_{11}^i & \theta_{12}^i & \theta_{13}^i\\
\theta_{21}^i & \theta_{22}^i & \theta_{23}^i
\end{array}
\right ]}, \label{T}
\end{equation}
which allows cropping, translation, scale and rotation to be applied to the input map.
We take the hidden state ${h_i}$ to calculate the parameters of transformation matrix ${T_i}$ with a fully-connected layers. Then we extract a region density map $r_i$ from the whole density map ${M_{i-1}}$ according to the transformation matrix ${T_i}$, which can be expressed as:
\begin{equation}
r_i=ST(M_{i-1}, T_i),
\end{equation}
where ST denotes the spatial transformer. The region map $r_i$ will be resized to a given size ${w*h}$ by bilinear interpolation.

\subsubsection{Region Density Map Refinement}
Inspired by the previous works[Sindagi and Patel, 2017a; 2017b] which have validated the importance of global contextual modeling in crowd counting, we also take into consideration the global context when performing attended density region refinement.
Intuitively, the global context contains the density level and density distribution of the given image.
We encode the global feature ${g}$ described in section~\ref{sec:coarse_map_generation} with two stacked fully-connected layers. The former one has 256 neurons and the latter one has $w*h$ neurons. We construct the global context map ${c_g}$ by reshaping the output of the last FC layer to a new tensor with a size of $w*h$.

\begin{figure}[t]
\centering
   \includegraphics[width=0.925\columnwidth]{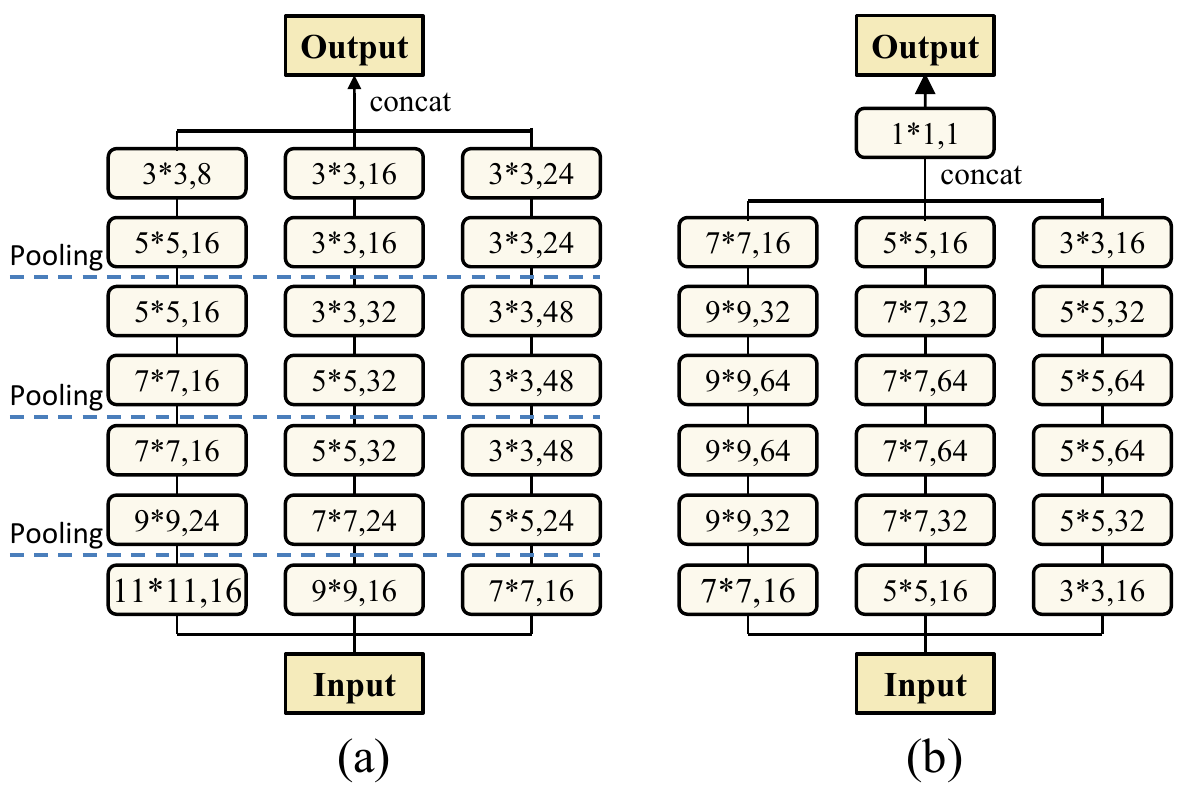}
\vspace{-3mm}
   \caption{(a) The architecture of Global Feature Embedding module for the global feature extraction. (b) The architecture of Local Refinement Network for the region density map refinement. The block with text $k*k, N$ denotes a convolutional layer with $k*k$ kernel size and $N$ output channels.
   }
\vspace{-3mm}
\label{fig:mcnn}
\end{figure}

After obtaining the region map $r_i$ and global context map ${c_g}$, we can perform crowd map refinement on the attentional region with a Local Refinement Network, which learns the residual between the region map $r_i$ and the ground truth of its corresponding region. As shown in Figure~\ref{fig:mcnn}(b), the Local Refinement Network also consists of three columns of CNNs. It takes the concatenation of $r_i$ and ${c_g}$ as the input and calculates the residual map of the attentional region.
Finally, we can get a new crowd density map ${M_i}$ by adding the inverse transformed residual map to the ${M_{i-1}}$, which can be described as:
\begin{equation}
M_i=M_{i-1} + IST(LRN(r_i, c_g), T_i^{-1}),
\end{equation}
\begin{equation}
T_i^{-1}= r*{
\left[ \begin{array}{ccc}
\theta_{22}^i & -\theta_{12}^i & \theta_{12}^i \theta_{23}^i -\theta_{13}^i \theta_{22}^i\\
-\theta_{21}^i & \theta_{11}^i & \theta_{13}^i \theta_{21}^i - \theta_{11}^i \theta_{23}^i
\end{array}
\right ]},
\end{equation}
\begin{equation}
r=\frac{1}{\theta_{11}^i \theta_{22}^i - \theta_{12}^i \theta_{21}^i},
\end{equation}
where IST is the inverse spatial transformer that can transform the residual map back to the attentional region according to the inverse of ${T_i}$, denoted as ${T_i^{-1}}$.

\subsection{Networks Optimization} ~\label{sec:optimization}
For the existence of fully-connected layers in our model, we need to ensure that all the input images are of the same resolutions.
Therefore, when optimizing the networks, we transform all training images or patches to the same resolutions and the details will be described in section~\ref{sec:experiments}.
Suppose there are N training images after data augmentation, and each image $I_i$ is annotated with a set of 2D points ${L_i}$ = \{${l_1,...,l_{C(i)}}$\}, where $C(i)$ is the total number of the labeled pedestrian in the image. We generate ground truth density map ${D_i}$ of image ${I_i}$ with the specific strategy introduced by Lempitsky et al.~\cite{lempitsky2010learning}, which ensures that the sum of the ground truth density map ${\sum_{p\in I_i} D_i(p)}$ over the entire image ${I_i}$  matches its crowd count with some negligible deviation.

Our networks are trained in an end-to-end manner and the overall loss function is defined as:
\begin{equation}
 Loss= \sum_{i=1}^{N} \| M_i^0 - D_i\|^2 + \sum_{i=1}^{N} \| M_i^n - D_i\|^2, \label{loss}
\end{equation}
where $M_i^0$ and ${M_i^n}$ are the initial map and the refined density map for image $I_i$. We adopt the TensorFlow ~\cite{abadi2016tensorflow} toolbox to implement our crowd counting network. The filter weights of all convolutional layers and fully-connected layers are initialized by truncated normal distribution with a deviation equal to 0.01. The learning rate is set to $ 10^{-4} $ initially and multiplied by 0.98 every 1K training iterations. The batch size is set to 1.
We optimize our networks parameters with Adam optimization~\cite{kingma2014adam} by minimizing the loss function Eq.(\ref{loss}).

\begin{table}
\newcommand{\tabincell}[2]{\begin{tabular}{@{}#1@{}}#2\end{tabular}}
  \centering
  \resizebox{8cm}{!} {
    \begin{tabular}{c|c|c|c|c}
    \hline
    \multirow{2}{*}{Method} &
    \multicolumn{2}{c|}{Part A} &
    \multicolumn{2}{c}{Part B} \\
    \cline{2-5}
    & MAE  & MSE & MAE & MSE \\
    \hline\hline
    ~\cite{zhang2015cross} & 181.8 & 277.7	& 32 & 49.8\\
    \hline
    ~\cite{zhang2016single} & 110.2 & 173.2 & 26.4 & 41.3\\
    \hline
    ~\cite{sindagi2017cnn} & 101.3 & 152.4 & 20 & 31.1\\
    \hline
    ~\cite{sam2017switching} & 90.4 & 135 & 21.6 & 33.4\\
    \hline
    ~\cite{sindagi2017generating} & 73.6 & 106.4 & 20.1 & 30.1\\
    \hline
    Ours & {\bf\textcolor{red}{69.3}} & {\bf\textcolor{red}{96.4}} & {\bf\textcolor{red}{11.1}} & {\bf\textcolor{red}{18.2}}\\
    \hline
    \end{tabular}
  }
  \vspace{-2mm}
   \caption{Performance evaluation of different methods on the ShanghaiTech dataset. Our proposed method outperforms the existing state-of-the-art methods on both parts of the ShanghaiTech dataset with a margin.
   }

  \label{tab:Result_ShanghaiTech}
\end{table}

\section{Experiments} ~\label{sec:experiments}
In this section, we first compare our method with recent state-of-the-art methods of the crowd counting task on four public challenging datasets. We further conduct extensive ablation studies to demonstrate the effectiveness of each component of our model.

\subsection{Evaluation Metric}
Following the existing work~\cite{zhang2016single,sindagi2017generating}, we adopt the mean absolute error (MAE) and mean squared error (MSE) as metrics to evaluate the accuracy of crowd counting estimation, which are defined as:
\begin{small}
\begin{equation}
   MAE = \frac{1}{N} \displaystyle\sum_{i=1}^{N} \|p_i - \hat{p}_i\|,
   MSE = \sqrt{\frac{1}{N} \displaystyle\sum_{i=1}^{N} \| p_i - \hat{p}_i \|^2},
\end{equation}
\end{small}
where ${N}$ is the total number of testing images, ${p_i}$ and ${\hat{p}_i}$ are the ground truth count and estimated count of the $i$-th image respectively. As discussed in section~\ref{sec:optimization}, ${\hat{p_i}}$ can be calculated by summing over the estimated crowd density map.

\subsection{Evaluations and Comparisons}
\begin{figure*}[t]
\begin{center}
 \includegraphics[width=1.75\columnwidth]{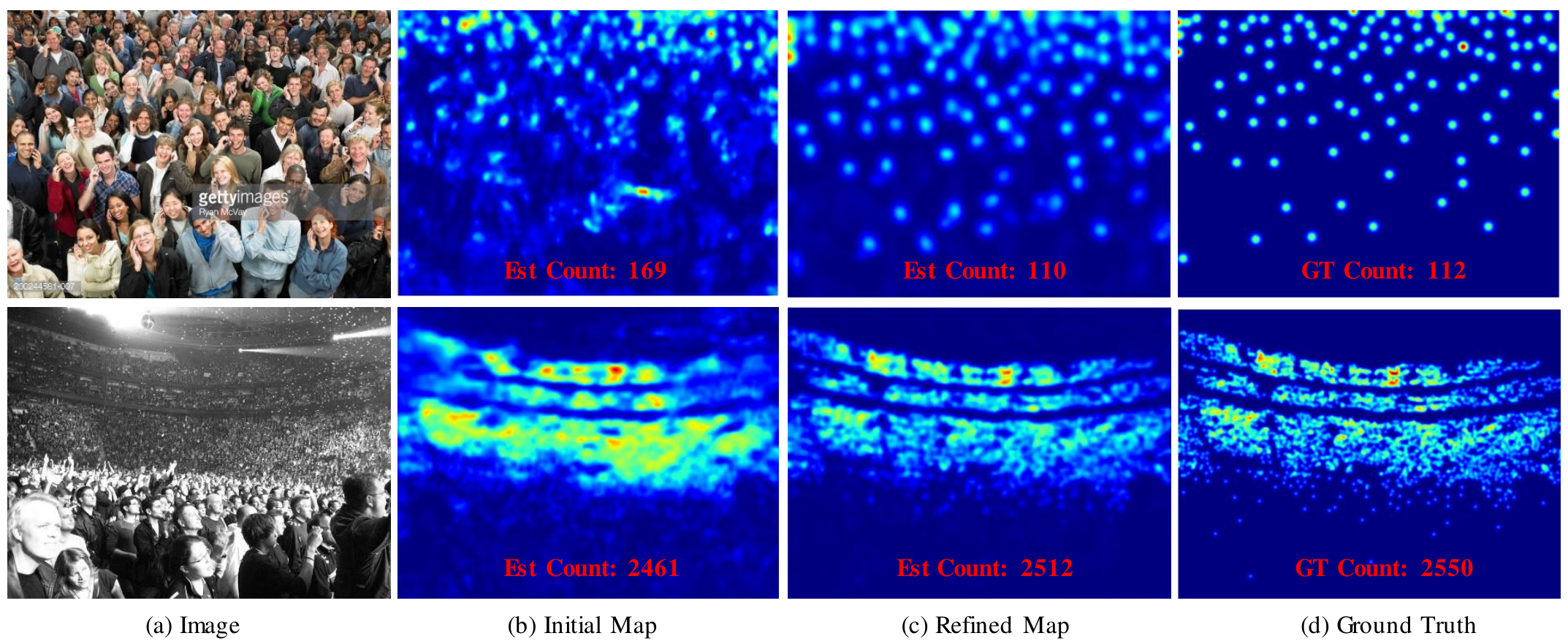}
 \vspace{-5mm}
\end{center}
   \caption{Visualization of the crowd density maps generated by our method on ShanghaiTech dataset. We can see that more accurate crowd density map and estimated count are obtained after refinement.
   }
\vspace{-3mm}
\label{fig:result_visual}
\end{figure*}

{\bf ShanghaiTech~\cite{zhang2016single}.} This dataset contains 1,198 images of unconstrained scenes with a total of 330,165 annotated people.
And it is divided into two parts: Part A with 482 images crawled from the Internet, and Part B with 716 images taken from the busy shopping streets. For each part, we resize the images to its maximal resolution when training and testing.

We compare our method with five recently published state-of-the-art methods on this dataset, i.e., Zhang et al.~\cite{zhang2015cross}, MCNN~\cite{zhang2016single}, Cascaded-MTL~\cite{sindagi2017cnn}, Switching-CNN~\cite{sam2017switching}, CP-CNN~\cite{sindagi2017generating}. As shown in Table~\ref{tab:Result_ShanghaiTech}, our proposed method outperforms other competing methods on both parts of the ShanghaiTech dataset. Specifically, our method achieves a significant improvement of 49.7\% in MAE and 39.5\% in MSE over the existing best-performing algorithm CP-CNN on Part B.

{\bf UCF\_CC\_50~\cite{idrees2013multi}.} As an extremely challenging benchmark, this dataset contains 50 annotated images of diverse scenes collected from the Internet. Except for different resolutions, aspect ratios and perspective distortions, the images of this dataset also suffer from a wide range of person counts, varying from 94 to 4,543.
Following the standard protocol discussed in ~\cite{idrees2013multi}, we split the dataset into five subsets and perform a five-fold cross-validation.
When training, we randomly crop some regions with a range of $[0.5, 0.9]$ from the original images and resize them to $1024*768$. The testing images are directly resized to the same resolution.

We perform comparison against several state-of-the-art methods based on deep-learning, including
Zhang et al.~\cite{zhang2015cross},
MCNN~\cite{zhang2016single},
Walach et al.~\cite{walach2016learning},
Cascaded-MTL~\cite{sindagi2017cnn},
Switching-CNN~\cite{sam2017switching},
CP-CNN~\cite{sindagi2017generating} and
ConvLSTM-nt~\cite{xiong2017spatiotemporal}.
As shown in Table~\ref{tab:Result-UCF}, our method achieves superior performance in comparison to other competing methods on UCF\_CC\_50 dataset.
Specifically, our method outperforms the existing best-performing method ConvLSTM-nt by 22.8\% over the MAE metric.

\begin{table}
\newcommand{\tabincell}[2]{\begin{tabular}{@{}#1@{}}#2\end{tabular}}
  \centering
    \begin{tabular}{c|c|c}
    \hline
    Method & MAE & MSE\\
    \hline
    \hline
    ~\cite{zhang2015cross}     & 467   & 498.5\\
    \hline
    ~\cite{zhang2016single}            &377.6  & 509.1\\
    \hline
    ~\cite{walach2016learning}    & 364.4 & 341.4\\
    \hline
    ~\cite{sindagi2017cnn}    & 322.8 & 341.4\\
    \hline
    ~\cite{sam2017switching}   & 318.1 & 439.2\\
    \hline
    ~\cite{sindagi2017generating}          & 295.8 & 320.9\\
    \hline
    ~\cite{xiong2017spatiotemporal}     & 284.5 & 297.1\\
    \hline
    Ours            & {\bf\textcolor{red}{219.2}} & {\bf\textcolor{red}{250.2}}\\
    \hline
    \end{tabular}
  \vspace{-2mm}
   \caption{Performance evaluation of different methods on the UCF\_CC\_50 dataset.
  }
  \vspace{-3mm}
  \label{tab:Result-UCF}
\end{table}

{\bf MALL~\cite{chen2012feature}.} This dataset was captured by a publicly accessible surveillance camera in a shopping mall with more challenging lighting conditions and glass surface reflections.
The video sequence consists of 2,000 frames with a resolution of ${320*240}$.
Following the same setting as~\cite{chen2012feature}, we use the first 800 frames for training and the remaining 1,200 frames for evaluation.
The images are kept the original resolution when training and testing.

We compare our method with Gaussian Process Regression~\cite{chan2008privacy}, Ridge Regression~\cite{chen2012feature}, Cumulative Attribute Regression~\cite{chen2013cumulative}, Count forest~\cite{pham2015count} and ConvLSTM~\cite{zhang2016single}. The results of all methods are summarized in Table~\ref{tab:Result-MALL}.
Our still-image based method outperforms the existing best method ConvLSTM, which utilized extra temporal dependencies from surveillance video to estimate the crowd count.

\begin{table}
\newcommand{\tabincell}[2]{\begin{tabular}{@{}#1@{}}#2\end{tabular}}
  \centering
    \begin{tabular}{c|c|c}
    \hline
    Method & MAE & MSE\\
    \hline
    \hline
    ~\cite{chan2008privacy}     & 3.72  & 20.1 \\
    \hline
    ~\cite{chen2012feature}                & 3.59  & 19.0 \\
    \hline
    ~\cite{chen2013cumulative} & 3.43  & 17.7 \\
    \hline
    ~\cite{pham2015count}                    & 2.50 & 10.0 \\
    \hline
    ~\cite{zhang2016single}                        & 2.24 & 8.5\\
    \hline
    Ours            & {\bf\textcolor{red}{1.72}} & {\bf\textcolor{red}{2.1}}\\
    \hline
    \end{tabular}
  \vspace{-3mm}
    \caption{Performance evaluation on the MALL dataset.
  }
  \label{tab:Result-MALL}
\end{table}

{\bf WorldExpo'10~\cite{zhang2015cross}.} This dataset contains 1,132 video sequences captured by 108 surveillance cameras during the Shanghai WorldExpo in 2010. The training set consists of 3,380 annotated frames from 103 scenes, while the testing images are extracted from other five different scenes with 120 frames per scene. The images are kept the original resolution for training and testing.
When testing, the Region of Interest (ROI) maps for each scene are provided and we only consider the crowd count under the ROI.

We compare our method with seven recently published state-of-the-art methods, including Chen et al.~\cite{chen2013cumulative}, Zhang et al.~\cite{zhang2015cross}, MCNN~\cite{zhang2016single}, Shang et al.~\cite{shang2016end}, Switching-CNN~\cite{sam2017switching}, CP-CNN~\cite{sindagi2017generating} and ConvLSTM~\cite{xiong2017spatiotemporal}.
As demonstrated in Table~\ref{tab:Result_WorldExpo}, our method achieves state-of-the-art performance with respect to the average MAE of five scenes. Specifically, our model gets the lowest MAE in Scene 2 and Scene 5, which are the two most challenging scenes in the testing set.
\begin{table*}
\centering
\newcommand{\tabincell}[2]{\begin{tabular}{@{}#1@{}}#2\end{tabular}}
  \centering
    \begin{tabular}{c|c|c|c|c|c|c}
    \hline
    Method & Scene1 & Scene2 & Scene3 & Scene4 & Scene5 & Average\\
    \hline\hline
    ~\cite{chen2013cumulative}      & {\bf\textcolor{red}{2.1}}	&   55.9 &	{\bf\textcolor{red} {9.6}}	 &  11.3 & {\bf\textcolor{red} {3.4}}  & 16.5\\
    \hline
    ~\cite{zhang2015cross}          & 9.8	&   {\bf\textcolor{blue}{14.1}} &  14.3 &  22.2 & 3.7	& 12.9\\
    \hline
    ~\cite{zhang2016single}         & 3.4	&   20.6 &	12.9 &	13	 & 8.1	& 11.6\\
    \hline
    ~\cite{shang2016end}           & 7.8	&   15.4 &  14.9 &	11.8 & 5.8	& 11.7\\
    \hline
    ~\cite{sam2017switching}        & 4.4	&   15.7 &	{\bf\textcolor{blue}{10}}	 &  11	 & 5.9	& 9.4\\
    \hline
    ~\cite{sindagi2017generating}   & 2.9	&   14.7 &	10.5 &	{\bf\textcolor{red} {10.4}} & 5.8	& {\bf\textcolor{blue} {8.86}}\\
    \hline
    ~\cite{xiong2017spatiotemporal} & 7.1	&   15.2 &	15.2 &	13.9 & {\bf\textcolor{blue}{3.5}}	& 10.9\\
    \hline
    Ours                            & {\bf\textcolor{blue}{2.6}}	&   {\bf\textcolor{red}{11.8}}  &	10.3 & {\bf\textcolor{red} {10.4}} & 3.7	& {\bf\textcolor{red} {7.76}} \\
    \hline
    \end{tabular}
  \vspace{-3mm}
  \caption{Mean Absolute Error of different methods on the WorldExpo’10 dataset. Our method achieves superior performance with respect to the average MAE of five scenes. The best results and the second best results are highlighted in {\textcolor{red}{red}} and {\textcolor{blue}{blue}}, respectively. Best viewed in color.}
  \vspace{-5mm}
  \label{tab:Result_WorldExpo}
\end{table*}

\begin{table}
      \centering
        \begin{tabular}{c|c|c|c|c}
        \hline
        \multirow{2}{*}{Method} &
        \multicolumn{2}{c|}{Part A} &
        \multicolumn{2}{c}{Part B} \\
        \cline{2-5}
        & MAE  & MSE & MAE & MSE \\
        \hline\hline
        Basic & 83.1 & 120.4 & 20.1 & 31.8\\
        \hline
        T & 75.9 & 109.2 & 16.5 & 21.9\\
        \hline
        T + S & 72.9 & 103.1 & 14.6 & 21.4\\
        \hline
        T + S + R & {\bf 69.3} & {\bf 96.4} & {\bf 11.6} & {\bf 19.5}\\
        \hline
        \end{tabular}
      \vspace{-2mm}
      \caption{Comparison of the performance of our model with different constraints of the spatial transformer on ShanghaiTech dataset. T, S, and R correspond to translation, scale, and rotation respectively.}
      \vspace{-3mm}
      \label{tab:components}
\end{table}

\subsection{Ablation Study}
In this section, we perform extensive ablation studies on ShanghaiTech dataset and demonstrate the effects of several components in our framework.

{\bf Effectiveness of Spatial Transformation:}
The transformation matrix defined in Eq.(\ref{T}) allows translation, rotation and scale to be applied to the input map.
To validate the effectiveness of each of them, we train our model with three different constraints of spatial transformer:
i) spatial transformation only with translation by directly setting $\theta_{11}$, $\theta_{22}$ to 1 and $\theta_{12}$, $\theta_{21}$ to 0, denoted as ${T}$;
ii) spatial transformation with translation and scale but without rotation by directly setting $\theta_{12}$, $\theta_{21}$ to 0, denoted as ${T+S}$;
iii) spatial transformation with translation, rotation and scale, denoted as ${T+S+R}$.
The performance of these variants are summarized in Table~\ref{tab:iteration} and the performance of initial density map is denoted as ``Basic''.
We can see that directly locating a region without scale and rotation operation for refinement can reduce the MAE from 83.1 to 75.9 on ShanghaiTech Part A.
When dynamically resizing the local regions, the MAE on Part A can be reduced to 73.2.
When simultaneously resizing and rotating the local regions, we can get the best performance on ShanghaiTech dataset.
The experiment has demonstrated the effectiveness of the spatial transformation for crowd density map refinement.

{\bf Effectiveness of Global Context:}
In order to validate the effectiveness of the global context, we train a variant of our model that refines the local region map without the global context.
As shown in Table~\ref{tab:context}, the performance would drop over all metrics when removing the global context.
Actually, the global context contains the information of density level and density distribution, which is beneficial for the crowd counting inference.

\begin{table}
      \centering
        \begin{tabular}{c|c|c|c|c}
        \hline
        \multirow{2}{*}{Method} &
        \multicolumn{2}{c|}{Part A} &
        \multicolumn{2}{c}{Part B} \\
        \cline{2-5}
        & MAE  & MSE & MAE & MSE \\
        \hline\hline
         W/O Global Context & 74.44 & 100.1 & 15.7 & 24.9\\
         \hline
         W/ Global Context & {\bf 69.3} & {\bf 96.4} & {\bf 11.6} & {\bf 19.5}\\
        \hline
        \end{tabular}
      \vspace{-2mm}
      \caption{Effectiveness verification of global context on the ShanghaiTech dataset.}
      \label{tab:context}
\end{table}

{\bf Effectiveness of Recurrent Refinement:} To validate the superiority of the recurrent refinement mechanism of our method, we train our model with different refinement iterations. As shown in Table~\ref{tab:iteration}, the performance gradually increases with more iterations and it slightly drops until 40 steps. Our method gets the best performance with 30 iterations, therefore we set the refinement iteration $n$ to 30 in our model for all datasets. Figure~\ref{fig:result_visual} shows the visual comparison with the refined maps and the initial maps and we can see that our method can generate more accurate crowd density maps after multi-step refinements.
\begin{table}
      \centering
        \begin{tabular}{c|c|c|c|c}
        \hline
        \multirow{2}{*}{Steps} &
        \multicolumn{2}{c|}{Part A} &
        \multicolumn{2}{c}{Part B} \\
        \cline{2-5}
        & MAE  & MSE & MAE & MSE \\
        \hline\hline
        $n$ = 0 & 83.1 & 120.4 & 20.1 & 31.8\\
        \hline
        $n$ = 10 & 82.1 & 111.3 & 18.8 & 30.4\\
        \hline
        $n$ = 20 & 73.2 & 113.1 & 11.2 & 19.3\\
        \hline
        $n$ = 30 & {\bf 69.3} & {\bf 96.4} & {\bf 11.1} & {\bf 18.2}\\
        \hline
        $n$ = 40 & 74.12 & 110.33 & 14.27 & 20.78\\
        \hline
        \end{tabular}
      \vspace{-2mm}
      \caption{ShanghaiTech dataset experimental results on the variants of our model using different refinement steps. Our method has the best performance when the density map is refined by $n=30$ steps.}
      \vspace{-4mm}
      \label{tab:iteration}
\end{table}

\section{Conclusion}
In this paper, we introduce a novel Deep Recurrent Spatial-Aware Network for crowd counting which simultaneously models the variations of crowd density as well as the pose changes in a unified learnable module. It can be regarded as a general framework for crowd map refinement. Extensive experiments on four challenging benchmarks show that our proposed method achieves superior performance in comparison to the existing state-of-the-art methods. In our future research, we plan to delve into the research of incorporating our model in other existing crowd flow prediction framework.

\section*{Acknowledgments}
This work was supported by State Key Development Program under Grant 2016YFB1001004, National Natural Science Foundation of China under Grant 61622214 and Grant 61702565, and Guangdong Natural Science Foundation Project for Research Teams under Grant 2017A030312006. This work was also sponsored by CCF-Tencent Open Research Fund. Wanli Ouyang is supported by the SenseTime Group Limited.

\bibliographystyle{named}
\bibliography{counting}

\end{document}